\documentclass{llncs}
\usepackage{graphicx}
\usepackage{subfig}
\usepackage{amsmath} 
\usepackage{amssymb}  
\usepackage{xspace}
\usepackage{ifthen}
\usepackage{xcolor}
\usepackage{enumerate}
\usepackage[pstricks1-10]{vaucanson-g}
\usepackage[free-standing-units=true]{siunitx}

\newcommand{\formatAction}[1]{{\textit{\textsf{#1}}}\xspace}
\newcommand{\formatFluent}[1]{{\small \emph{#1}}\xspace}

\newcommand{\go}[2]{\formatAction{go.#1.#2}}
\newcommand{\arrived}[2]{\formatAction{at.#1.#2}}
\newcommand{\alerton}{\formatAction{alert.on}}
\newcommand{\alertoff}{\formatAction{alert.off}}

\newcommand{\set}[1]{\{#1\}}
\newcommand{\gr}{{GR(1)}\xspace}

\newcommand{\G}{\square}
\newcommand{\F}{\Diamond}
\newcommand{\X}{\circ}
\newcommand{\WaitFor}{\mathbf{\mathcal{W}}}
\newcommand{\W}{\WaitFor}
\renewcommand{\implies}{\ensuremath{\Rightarrow}}

\DeclareSIUnit{\rpm}{rpm}
\newcommand{\arch}[1]{{\small \textsc{#1}}\xspace}

\newcommand{\comments}[1]{}

\newboolean{showcomments}
\setboolean{showcomments}{true} 
\ifthenelse{\boolean{showcomments}}{
	\newcommand{\nbc}[3]{
		{\colorbox{#3}{\bfseries\sffamily\scriptsize\textcolor{white}{#1}}}
		{\textcolor{#3}{\sf\small$\langle$\textit{#2}$\rangle$}}}
}{
	\newcommand{\nbc}[3]{}
	
}

\begin{document}

\title{Hybrid Control from Scratch: 
A Design Methodology for Assured Robotic Missions}
%
%
\author{Tom\'as Liendro \and Sebasti\'an Zudaire}
%
%
%
\institute{Instituto Balseiro, Universidad Nacional de Cuyo,\\ R\'io Negro, 
Argentina,\\
\email{\{tomas.liendro, sebastian.zudaire\}@ib.edu.ar}}

\maketitle              

\begin{abstract}
Robotic research over the last decades have lead us to different architectures to 
automatically synthesise discrete event controllers and implement these motion and 
task plans in real-world robot scenarios. However, these architectures usually build on
existing robot hardware, generating as a result solutions that are influenced and/or 
restricted in their design by the available capabilities and sensors. In contrast to 
these approaches, we propose a design methodology that, given a specific domain of 
application, allowed us to build the first end-to-end implementation of an autonomous 
robot system that uses discrete event controller synthesis to generate assured mission 
plans. We validate this robot system in several missions of 
our target domain of application.
\keywords{discrete event control, hybrid control, cyber-physical systems, 
design 
methodology}
\end{abstract}

\section{Introduction}

Hybrid controllers are gaining increasing attention as a 
way to translate discrete mission specifications into continuous motion of 
robots~\cite{Gazit09}. This is achieved through planning techniques that allow us to 
synthesise correct-by-construction discrete controllers from these 
specifications~\cite{piterman06} using an adequate discrete 
abstraction for the robot's capabilities and workspace. Then, a hybrid control 
layer~\cite{Gazit08} is responsible of interpreting these plans into a set of inputs for a 
lower control level that commands the robot's motion and other functionalities. The 
result is an autonomous cyber-physical system guaranteed to satisfy the original 
specifications if 
a set of assumptions holds.

Hybrid controllers are possible thanks to contributions from the fields of 
Robotic, Control and Automated Software Engineering. On one hand, the robotic and control 
community studies the control and automation of continuous and discrete variable dynamic 
systems, developing design and implementation techniques for motion and control of robot 
systems~\cite{twowheel4}. On the other hand, the automated software community has 
developed many complex automated reasoning tools to automatically generate plans from 
specifications expressed in formal logic languages like Linear Temporal Logic 
(LTL)~\cite{Belta07,LTLMoP}.

There has been extensive research on different architectures for hybrid controllers and 
their implementation on functioning autonomous systems~\cite{Jing17,Atlas}. In 
this work we focus on the inverse problem: if one were to design a robot system to 
implement a hybrid controller for a set of mission patterns, what would this design and 
implementation be? 

In order to help answer this question, the main contributions of this work are: (1) a 
design methodology for developing assured robotic missions, (2) the first end-to-end 
design and implementation of an autonomous robot system that uses discrete event 
controller synthesis to generate assured mission plans, using the proposed methodology for 
a specific surveillance scenario. 

For this, we build on a corpus of knowledge 
ranging from higher level planning problems to the lower level 
mechanical and  hardware design, effectively building on all levels of abstraction. We 
show our hybrid control architecture and design methodology 
in Sections~\ref{sec:component} and~\ref{sec:methodology}, and use it to build a
surveillance robot (Section~\ref{sec:construction}) that we validate 
in several mission scenarios (Section~\ref{sec:validation}).


\section{Background} \label{sec:background}


We present in this section the basic formalism for the controller synthesis technique we 
used and basic definitions of the rest of the concepts we work with. For more 
insight into these concepts please refer to~\cite{Nahabedian18} and~\cite{franklin}.

\textbf{Labelled Transition Systems (LTS).}
The dynamics of the interaction of a robot with its environment will be modelled using 
LTS~\cite{Keller76}, which are automata where
transitions are labelled with events that constitute the interactions
of the modelled system with its environment.
We partition events into controlled and uncontrolled to specify
assumptions about the environment and safety requirements for a
controller.
Complex models can be constructed by LTS composition. We use a
standard definition of \textit{parallel composition} ($\|$) that models the
asynchronous execution of LTS, interleaving non-shared actions and
forcing synchronisation of shared actions.

\textbf{Fluent Linear Temporal Logic (FLTL).}
In order to describe environment assumptions and system goals it is
common to use formal languages like FLTL~\cite{gianna03}, a variant
of linear-time temporal logic that uses fluents to describe states
over sequences of actions.
A \emph{fluent} \formatFluent{fl} is defined by  a set of events that make it 
true ($Set_\top$), a set of events 
that make it false ($Set_\bot$) and an initial value ($v$) true ($\top$) or false 
($\bot$): $\formatFluent{fl} = \langle Set_\top, Set_\bot, v \rangle$. 
We may omit set notation for singletons and use an action label 
\formatAction{\ensuremath{\ell}}
for the fluent defined as
$\formatFluent{fl} = \langle \formatAction{\ensuremath{\ell}}, 
\emph{Act}\setminus\set{\formatAction{\ensuremath{\ell}}},\bot
\rangle$. Thus, the fluent \formatAction{\ensuremath{\ell}} is only true just after the
occurrence of the action \formatAction{\ensuremath{\ell}}.

FLTL is defined similarly to propositional LTL but where a fluent
holds at a position $i$ in a trace $\pi$ based on the events occurring
in $\pi$ up to $i$. Temporal connectives are interpreted as usual:
$\F \varphi$, $\G \varphi$, and $ \varphi \W \psi$ mean that $\varphi$
eventually holds, always holds, and (weakly) holds until $\psi$, respectively.

\textbf{Discrete Event Controller Synthesis.}
Given an LTS $E$ with a set of controllable actions $L$ and a task specification $G$ 
expressed in FLTL, the goal of controller synthesis is
to find an LTS $C$ such that $E\|C$: (1) is deadlock free, (2) $C$
does not block any non-controlled actions, and (3) 
every trace of
$E\|C$ 
satisfies $G$. We say that a control problem $\langle E, G, L \rangle$ is 
\textit{realizable} if 
such an LTS 
$C$ exists.
The tractability of the controller synthesis depends on the size of the problem (i.e. 
states of $E$ and size of $G$) and also on the fragment of the logic used for $G$.
When goals are restricted to  \gr the control problem can be solved in polynomial
time~\cite{piterman06}. \gr formulas are of the form $\bigwedge_{i=1}^n \G\F \psi_i 
\implies 
\bigwedge_{i=1}^m \G\F \varphi_i$  where $\psi_i$ and $\varphi_i$
are Boolean combinations of fluents 
that refer to assumptions and goals, respectively.
In this paper we use MTSA~\cite{MTSA} for solving control problems.

\begin{figure}[bt]
\centering
\vspace{-6mm}
\hfill
  \subfloat[Dynamic system]{
	\begin{minipage}[c][0.45\width]{
	   0.3\textwidth}
	   \centering
	  \includegraphics[width=\linewidth]{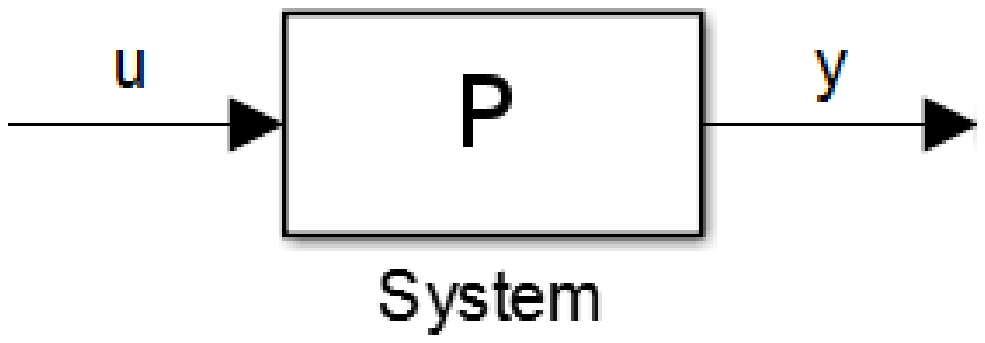}
	  \vspace{-6mm}
	  \label{fig:system}
	\end{minipage}}
 \hfill 
  \subfloat[Feedback-controller scheme]{
	\begin{minipage}[c][0.25\width]{
	   0.55\textwidth}
	   \centering
	\includegraphics[width=\linewidth]{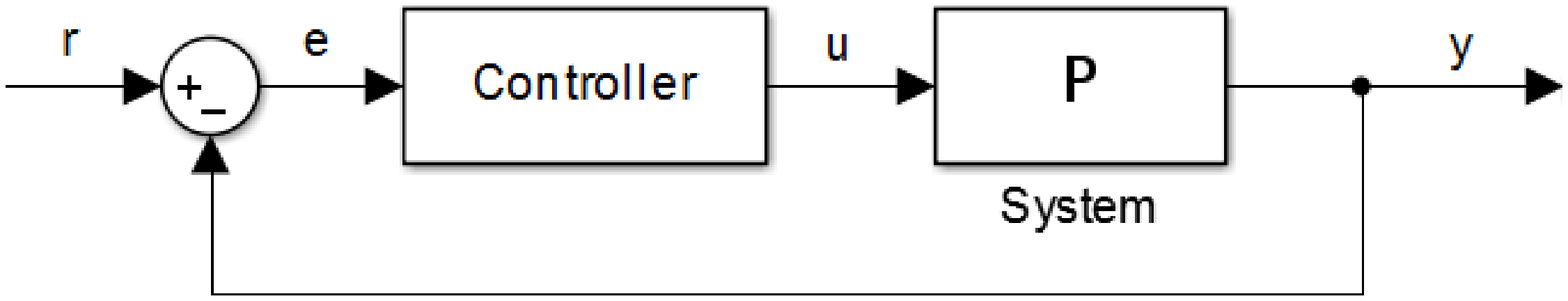}
	  \vspace{-6mm}
	  \label{fig:feedback}
	\end{minipage}}
\hspace{6mm}
\caption{Continuous variable dynamic systems abstracted to design controllers.}
\end{figure}

\textbf{Feedback-control.} Continuous variable dynamic systems (e.g., a robot wheel) can 
usually be abstracted into a system $P$ with a set of inputs $\vec{u}$ (e.g., voltage 
to the motor driving the wheel) and a set of outputs $\vec{y}$ (e.g., angular speed of 
the wheel) as shown in Fig.~\ref{fig:system}. The typical control problem in this 
setting is defined as finding the set of inputs $\vec{u}$ that drives the system $P$ to a 
certain desired configuration of the output $\vec{y}$, both in the final value of the 
variables as well as in the transition period, which can be approached in a open-loop or 
closed-loop form.
Open-loop control consists of generating the set of inputs for the system $P$
from a model of the system's behaviour. This approach doesn't take into account errors in 
the model or external perturbations to the system. 
A more robust approach is closed-loop control (i.e., \emph{feedback-control}), which uses 
direct or filtered measurements of the output $\vec{y}$ to compute the input $\vec{u}$ 
that helps minimise the error $\vec{e} = \vec{r} - \vec{y}$ (see 
Fig.~\ref{fig:feedback}), according to a set of specifications.


\textbf{Hybrid Controller.} In robotics, the difference in the continuous 
vs. discrete description of the real world, and in the 
interaction between \emph{discrete event controllers} and \emph{feedback-controllers} (or 
other robot actuators and sensors) require a non-trivial translation task that is 
implemented in a \emph{hybrid control layer} \cite{Belta07,Fainekos05}. 

To implement these plans, the hybrid control layer is responsible of taking high-level 
commands (e.g., the \formatAction{camera.on} action from~\cite{Gazit09}) and generating 
low-level inputs for the robot (e.g., a low-level voltage pulse train to trigger the 
camera into capture mode). When this translation is specifically related to motion, there 
is extensive work in the area of 
\emph{motion planning} (e.g.,~\cite{Belta04,Conner03,Castro15}).
Translating motion commands into continuous movement generally 
involves obtaining a set of discrete inputs that serve as a reference signal $\vec{r}$ 
for the \emph{feedback-controllers}, which then produce controlled 
movement of the robot~\cite{Fainekos05}.

\section{Related Work} \label{sec:related}

The design of hybrid control architectures for end-to-end implementations, i.e., that 
integrate components ranging from planning to robot hardware, has been 
approached by many (e.g., \cite{Belta07,Fainekos05,Gazit09,Gazit08,Jing17,Atlas}).
However, this process usually begins with partially or fully 
functional low-level robot hardware and software (e.g., \cite{Jing17}). In these 
scenarios the design problem consists in finding a discrete abstraction for the 
planning layer that captures the 
robots capabilities and workspace~\cite{Atlas}, and that is compatible with the desired 
domain of application. The key challenge is that the state space of the chosen 
abstractions must not be too big so as lead to intractability of the synthesis problem 
(e.g., increasing number of discrete locations leads to polynomial growth in synthesis  
time~\cite{Dathathri17}), nor too small so as to fail to capture important aspects from 
the continuous environment of the robot (e.g., the abstraction problems presented 
in~\cite{Castro15}). 

To the best of our knowledge, our work is the first to tackle the design of hybrid 
controllers as a whole, with the distinctive feature of building a custom robot that 
accommodates to various choices made during the design process.

\section{Component Architecture} \label{sec:component}

\begin{figure}[bt]
 \centering
 \includegraphics[width=0.8\linewidth]{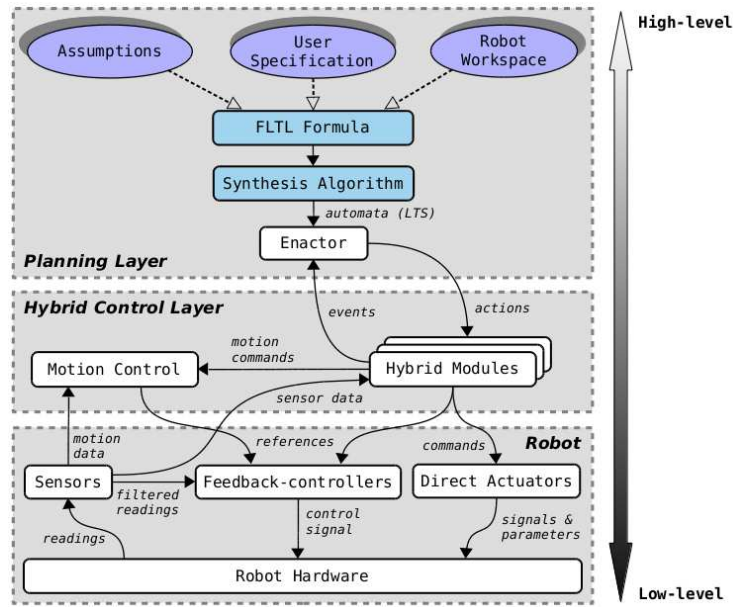}
 \caption{Hybrid control architecture showing interaction between components.}
 \label{fig:architecture}
\end{figure}

To approach an end-to-end design, we must first define a 
component architecture to identify the different elements of the system and their 
interaction. We build on several architectures taken from the control~\cite{Yong06}, 
robotic~\cite{Gazit08} and software communities~\cite{MORPH15} to define the 
architecture shown in Fig.~\ref{fig:architecture}, where arrows indicate interaction 
between components. We will describe this architecture using an
\arch{alternative font} when referring to elements in Fig.~\ref{fig:architecture}.

\textbf{Planning Layer}: This layer uses as an input high-level 
abstractions of the environment in which the robot moves and interacts. This input can be 
divided into a \arch{Robot Workspace} that provides a discretized notion of the 
capabilities of the robot (see Section~\ref{sec:abstraction}), a set of 
\arch{Assumptions} about the robot's environment and capabilities (e.g., a 
\formatAction{camera.on} action can only be followed by a \formatAction{camera.off}) and 
a \arch{User Specification} that captures what the mission for the robot is. All these 
elements can be encapsulated into a \arch{FLTL Formula} and then use a 
\arch{Synthesis Algorithm} to produce a correct-by-construction discrete event 
controller guaranteed to satisfy mission, which is represented as an \arch{automata} in 
LTS.

A component is then required to 
interpret the plan encoded in its LTS representation. The \arch{Enactor} consists of an 
interpretation module that knows which module to call from the available \arch{Hybrid 
Modules} each time controllable \arch{actions} are enabled in the plan, and continuously 
listens for discrete \arch{events} that may be triggered during the mission. 

\textbf{Hybrid modules}: These modules provide the main translation process 
between the continuous and discrete representations of the environment. To 
implement certain discrete \arch{actions}, references may be trivially computed at this 
level and sent to the \arch{Feedback-controllers} (e.g., to implement a 
\formatAction{drill.fast} action, the reference angular speed of a feedback-controller 
controlling the drilling speed may be set to \SI{3000}{\rpm}). For other 
actions, the \arch{Hybrid Modules} may command a \arch{Direct Actuator} (e.g., 
turning on a led). The process becomes more complicated for motion capabilities, so a 
\arch{Motion Control} or \emph{motion planner} module may be used to provide additional 
computation from the available \arch{motion data} to generate a set of \arch{references} 
for the \arch{Feedback-controllers}. From the available \arch{sensor data}, discrete 
\arch{events} may be generated.

\textbf{Robot}: This layer includes the physical \arch{Robot Hardware} as well as 
modules that provide the basic capabilities of the robot through 
\arch{Feedback-controllers} and \arch{Direct Actuators}. The difference between these two 
is that due to its closed-loop nature, the first require \arch{filtered readings} from 
the \arch{Sensors} to generate \arch{control signals}.

\section{Design Methodology} \label{sec:methodology}

In Fig.~\ref{fig:design} we show our methodology through the system design and 
implementation. Comparing with Fig.~\ref{fig:architecture}, the design flows from 
the highest levels of abstraction to the lowest, but the final component to be 
implemented is the \arch{Hybrid Control Layer}. Next we describe each stage of the 
process 
with more detail:

\begin{figure}[bt]
 \centering
 \includegraphics[width=1\linewidth]{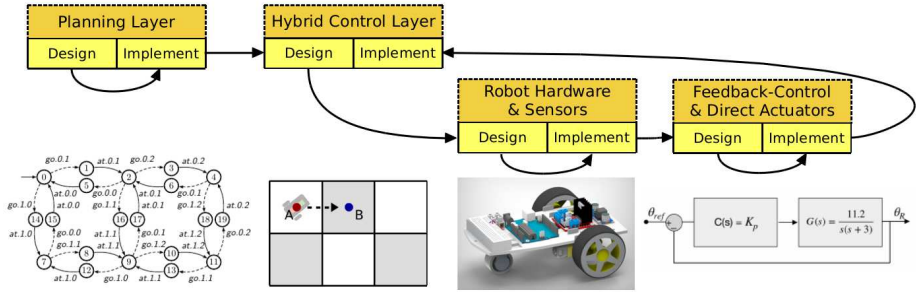}
 \caption{Design methodology. A representative image for 
each stage is shown.}
 \label{fig:design}
\end{figure}

\textbf{Planning Layer}: Our design methodology is guided by the desired domain 
of application, this means that we must first ensure that, for the missions we wish to 
specify and run, we have a compatible 
discrete abstraction and tractable synthesis algorithm. For example, common patrol or 
surveillance missions~\cite{Castro15,Menghi19} 
may be reasonably 
abstracted using grid-based discretization of the movement region of the robot, i.e., 
workspace. Multiple synthesis algorithms have shown tractability for this kind of 
specifications and abstractions (e.g.,~\cite{LTLMoP,Wolff13ICRA}), meaning that this is a 
reasonable approach for this family of missions. At this stage we both design the 
\arch{Planning Layer} and implement it by synthesising missions and eventually enacting 
them in dummy environments.

\textbf{Hybrid Control Layer (design)}: When the discrete abstraction is defined, 
we must provide a hybrid controller and robot hardware capable of implementing it. 
Common approaches usually involve using available working robot hardware on top of which 
they define a hybrid control architecture (e.g.,~\cite{Jing17}), but has the 
downside of incurring in complex architectures and components as a result. For instance, 
if we build on a common car-like robot, movement from one point to another will require at 
the very least the computation of a Dubins path or Reeds-Shepp 
path~\cite{ReedsShepp}. For certain space discretizations, this may also require 
sophisticated resynthesisation loops to find feasible trajectories (see 
~\cite{Castro15}). All this added complexity could have been avoided by selecting more 
adequate hardware for movement in a 2D plane such as omnidirectional wheels~\cite{Wen17}.

Our hypothesis is that many simplifications can be made by continuing the design process 
at the hybrid control layer. Designing the robot motion capabilities will require first 
designing a motion planning or motion control strategy to move the robot according to the 
discrete abstraction used. The same goes for other capabilities that require 
controlled command of the actuators.

\textbf{Robot Hardware \& Sensors}: Once reasonable motion and actuator strategies have 
been designed for the hybrid control layer, only the bottom low-level robot-related 
components are left. Robot hardware and sensors must be selected to comply with the design 
requirements defined during the previous stage.

\textbf{Feedback-Control \& Direct Actuators}: When the previous stage is 
implemented, the design of feedback-controllers and other actuators can commence, 
since they both require a real robot in which to validate their control algorithms.

\textbf{Hybrid Control Layer (implement)}: The final stage of the methodology 
involves integrating all the components into a fully working hybrid controller. For this, 
several \arch{Hybrid Modules} must be implemented as well as the \arch{Motion Control} 
strategy designed earlier.

\section{Example Use of the Design Methodology} \label{sec:construction}

In this section we use the design methodology described in Section~\ref{sec:methodology} 
to build a robot system for a surveillance mission scenario.

\subsection{Planning Layer} \label{sec:abstraction}

We first have to define our domain of application, i.e., the missions we want to 
do. The work in~\cite{Menghi19} provides common mission specification 
patterns found in the robotic literature, from which we aim to achieve all the \emph{Core 
Movement Patterns} (visiting and patrolling), \emph{Avoidance} 
patterns (prohibited regions) and the \emph{Reaction} patterns (e.g., turning on a camera 
when inside a region). We will work with 
low duration missions ($<$ \SI{30}{\minute}) on a small 2D plane (approx. 
$\SI{2}{\metre} \times \SI{3}{\metre}$) and restrict the reaction patterns into an alert 
with sound and leds.

\begin{figure}[bt]
\centering
\hfill
  \subfloat[]{
	\begin{minipage}[c][1.5\width]{
	   0.2\textwidth}
	   \centering
	  \includegraphics[width=0.9\linewidth]{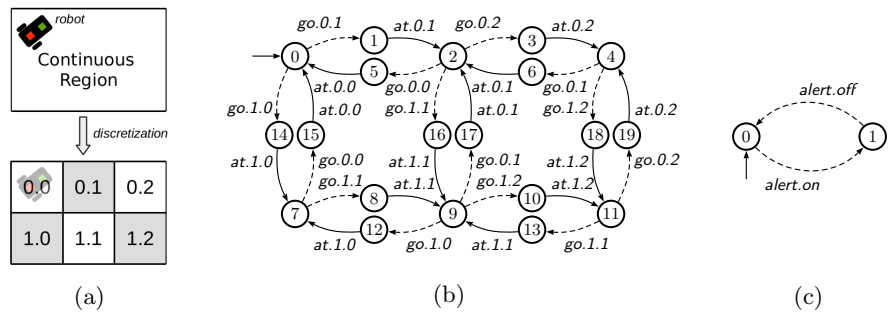}
	  \label{fig:disc}
	\end{minipage}}
 \hfill 
  \subfloat[]{
	\begin{minipage}[c][0.54\width]{
	   0.55\textwidth}
	   \centering
	\input{src/lts/discretization}
	  \label{fig:LTSdisc}
	\end{minipage}}
 \hfill 
  \subfloat[]{
	\begin{minipage}[c][1.5\width]{
	   0.2\textwidth}
	   \centering
	\input{src/lts/alert}
	  \label{fig:LTSalert}
	\end{minipage}}
\caption{(a) Grid-based discretization of a continuous region, (b) LTS for 
the movement capabilities of the robot, (c) LTS for the alert capability. \textit{Note:} 
dashed lines are controllable actions while continuous lines denote uncontrollable 
events.}
\end{figure}

For surveillance and visiting missions, its common to use space 
discretization of the robot's workspace~\cite{Belta07}. 
As the robot will be performing its tasks in an rectangular area, it 
is reasonable to discretize the robot's working area by using an uniform rectangular grid 
containing $N \times M$ cells as in~\cite{Wolff13ICRA}.
The minimum size of the cells is limited by the ability of the robot to move fully 
contained in it while manoeuvring to the next~\cite{Belta04}. This allows us to avoid 
discrete regions 
by specifying at the planning level never to take transitions that lead to prohibited 
regions. Furthermore, we must only allow movement to adjacent cells, since 
moving in diagonal may lead to entering forbidden cells along the way.

Fig.~\ref{fig:disc} and~\ref{fig:LTSdisc} show how we derive from the 
grid-based discretization of a region a LTS modelling the motion behaviour of the robot. 
We abstract motion into two actions: a controllable \go{i}{j} action to command the robot 
to move to the discrete location $(i,j)$ of the grid, and the respective 
uncontrollable \arrived{i}{j} event to indicate that the robot arrived correctly at this 
location. This LTS also includes certain assumptions about the 
movement of the robot: a \go{1}{2} action can only be followed by a 
\arrived{1}{2} event and between \go{i}{j} and \arrived{i}{j} no other 
\formatAction{go} 
can be issued. 
For the alert capability we may use a simple LTS abstraction as shown in 
Fig.~\ref{fig:LTSalert}, where each controllable 
action is modelled as instantaneous. The environment $E$ consists of the parallel 
composition of these two LTS.

We will use FLTL formulae to specify missions. For example, a patrol mission between the 
discrete locations $(0,0)$ and $(1,2)$, where the robot must turn on its alert in 
the bottom row locations and turn it off in the top row, and always avoid location 
$(0,2)$, can be specified with the fluents defined in (\ref{eqn:fl_ex}) and the 
formula $\varphi=(\varphi_1 \wedge \varphi_2 \wedge \varphi_3)$. 
This formula relates to the 
\emph{Patrolling} ($\varphi_1$), \emph{Global avoidance} ($\varphi_2$) and \emph{Prompt 
Reaction} ($\varphi_3$) patterns in~\cite{Menghi19}.
\begin{equation} \label{eqn:fl_ex}
 \begin{aligned}
  \formatFluent{AtBot} &= \langle \set{\arrived{1}{0}, \arrived{1}{1}, 
\arrived{1}{1}}, \set{\go{0}{0}, \go{0}{1}, \go{0}{2}}, \bot \rangle \\
  \formatFluent{AtTop} &= \langle \set{\arrived{0}{0}, \arrived{0}{1}, 
\arrived{0}{2}}, \set{\go{1}{0}, \go{1}{1}, \go{1}{2}}, \top \rangle \\
  \formatFluent{Alert} &= \langle \alerton, \alertoff, 
\bot \rangle
 \end{aligned}
\end{equation}
\begin{equation} \label{eqn:sp_ex}
 \begin{aligned}
\varphi_1 &= (\G\F \arrived{0}{0}) \wedge (\G \F \arrived{1}{2}) \\
\varphi_2 &= (\G\neg \arrived{0}{2}) \\
\varphi_3 = \G \Big(\big((\formatFluent{AtBot} \wedge \neg 
\formatFluent{Alert})&\implies \X 
\alerton\big) \wedge \big((\formatFluent{AtTop} \wedge \formatFluent{Alert}) \implies 
\X \alertoff\big) \Big)
 \end{aligned}
\end{equation}

We used the MTSA tool to solve the control problem derived from the environment 
$E$ and the above specifications. 
This tool proved tractability on low-end hardware (see Section~\ref{sec:hardware}) for 
hundreds of discrete locations.
We use the obstacle 
sensor described in Section~\ref{sec:hardware} to automatically specify locations 
to avoid ($\varphi_2$).

An \arch{Enactor} for these plans was implemented by parsing the output discrete event 
controllers generated with MTSA into an equivalent data structure in Python to be able to 
integrate with the modules in the hybrid layer. A queue mechanism 
allows it to run as an independent process, calling the controllable actions of the plan 
(implemented in the \arch{Hybrid Modules}) as they are enabled and immediately processes 
any uncontrollable event that may be generated.

%
%
%

\subsection{Hybrid Control Layer Design} \label{sec:hybrid}

Our target domain of application requires developing movement and non-movement (i.e., 
alert capability) functionality for the robot. Implementing an alert capability is 
pretty straight forward for simple temporised alarms. However, commanding movement of 
the robot is not, so we will focus on its design process.

\begin{figure}[bt]
\centering
\hfill
  \subfloat[]{
	\begin{minipage}[c][0.6\width]{
	   0.3\textwidth}
	   \centering
	  \includegraphics[width=0.9\linewidth]{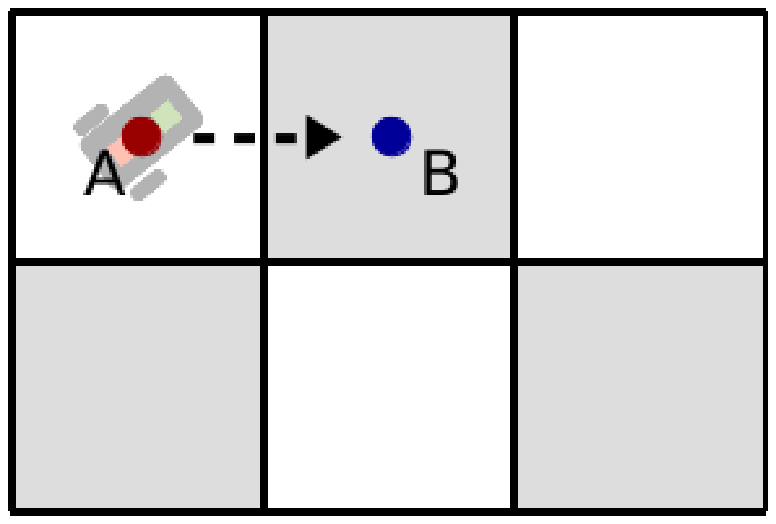}
	  \label{fig:controlprob}
	\end{minipage}}
 \hfill 
  \subfloat[]{
	\begin{minipage}[c][0.6\width]{
	   0.3\textwidth}
	   \centering
	  \includegraphics[width=0.9\linewidth]{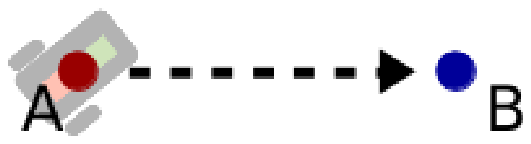}
	  \label{fig:motionplanning}
	\end{minipage}}
 \hfill 
  \subfloat[]{
	\begin{minipage}[c][0.6\width]{
	   0.3\textwidth}
	   \centering
	  \includegraphics[width=0.9\linewidth]{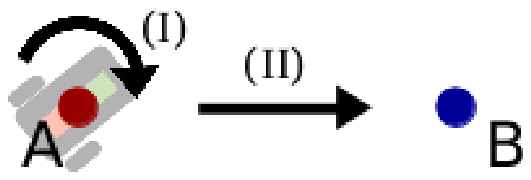}
	  \label{fig:twostep}
	\end{minipage}}
 \hspace{6mm}
\caption{(a) Adjacent movement capability of the robot, (b) Abstracted motion planning 
problem, (c) Two-step solution to the problem.}
\end{figure}

Given the discrete abstraction we defined in Section~\ref{sec:abstraction} that handles at 
the higher level static-obstacle avoidance, our motion planning problem reduces to finding 
a set of control inputs for the robot to go from one discrete location to an adjacent one. 
This simple problem can be solved without relying on sophisticated motion planners 
(e.g., RRT~\cite{Carpin02}) using either omnidirectional robots or with a motion control 
solution over an unidirectional moving robot. The first solution (e.g.,~\cite{Wen17}) 
just pushes the complexity and cost of the problem to a lower level. Here we show that a 
carefully selected simple motion control strategy can simplify the design and 
implementation process of the lower mechanical and hardware level, as stated in 
Section~\ref{sec:methodology}.

The motion problem is illustrated in Fig.~\ref{fig:controlprob}, were we see 
the robot wanting to do a transition through a \formatAction{go} action to an adjacent 
cell. The abstracted motion planning problem is shown in 
Fig.~\ref{fig:motionplanning} and a very simple solution to this problem is presented in 
Fig.~\ref{fig:twostep} for an unidirectional moving robot:

\begin{enumerate}[(I)]
 \item Rotate until the ``front'' of the robot points in 
the direction of the target.
 \item Move forward along this direction until the 
target is reached.
\end{enumerate}

This type of motion control solution requires both a robot with a unidirectional forward 
moving capability and the functionality of rotating around its centre axis (i.e., without 
displacement of its centre of mass), which can be easily satisfied and implemented with 
a low-cost two-wheel robot as in~\cite{twowheel4,twowheel2,twowheeljaiio}.

\subsection{Robot Hardware and Sensors} \label{sec:hardware}

\begin{figure}[bt]
\centering
\hfill
  \subfloat[]{
	\begin{minipage}[c][0.4\width]{
	   0.65\textwidth}
	   \centering
	  \includegraphics[width=1\linewidth]{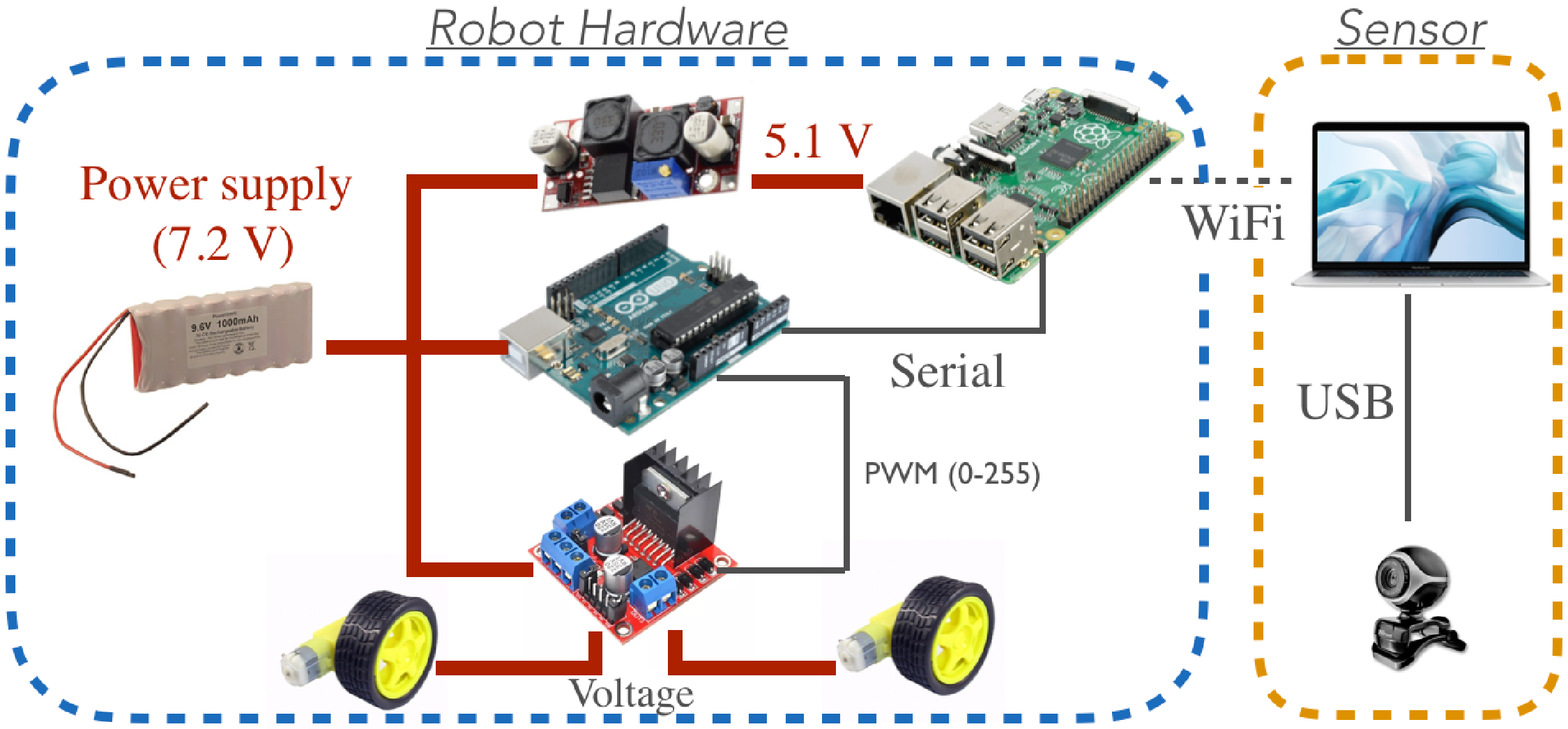}
	  \label{fig:conexion}
	\end{minipage}}
 \hfill 
  \subfloat[]{
	\begin{minipage}[c][0.8\width]{
	   0.33\textwidth}
	   \centering
	  \includegraphics[width=1\linewidth]{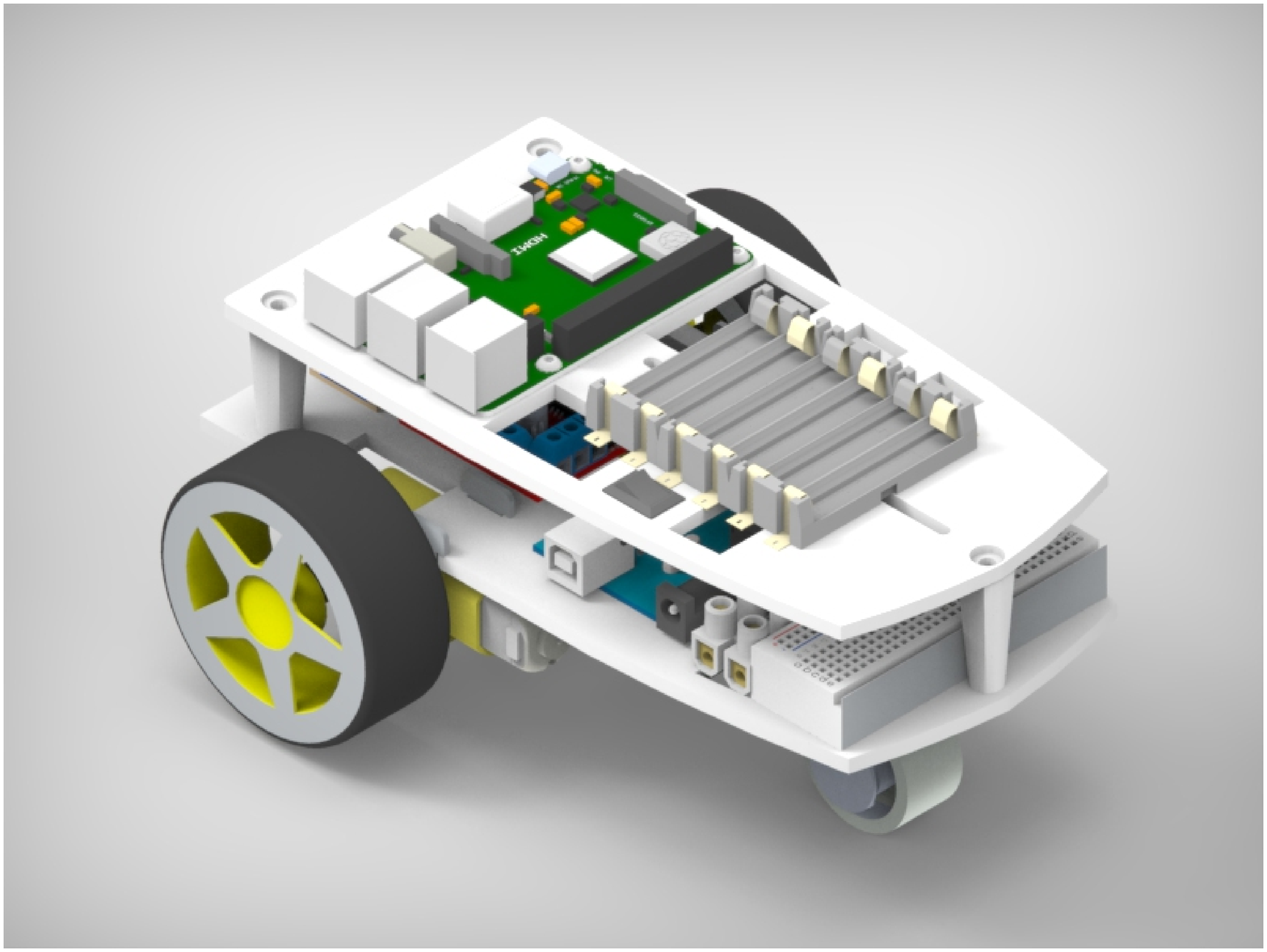}
	  \label{fig:render}
	\end{minipage}}
 \hfill
\caption{(a) Hardware components and connection scheme, (b) Main platform of the robot 
showing the component layout.}
\end{figure}

\begin{figure}[bt]
\centering
\hfill
  \subfloat[]{
	\begin{minipage}[c][0.56\width]{
	   0.65\textwidth}
	   \centering
	  \includegraphics[width=1\linewidth]{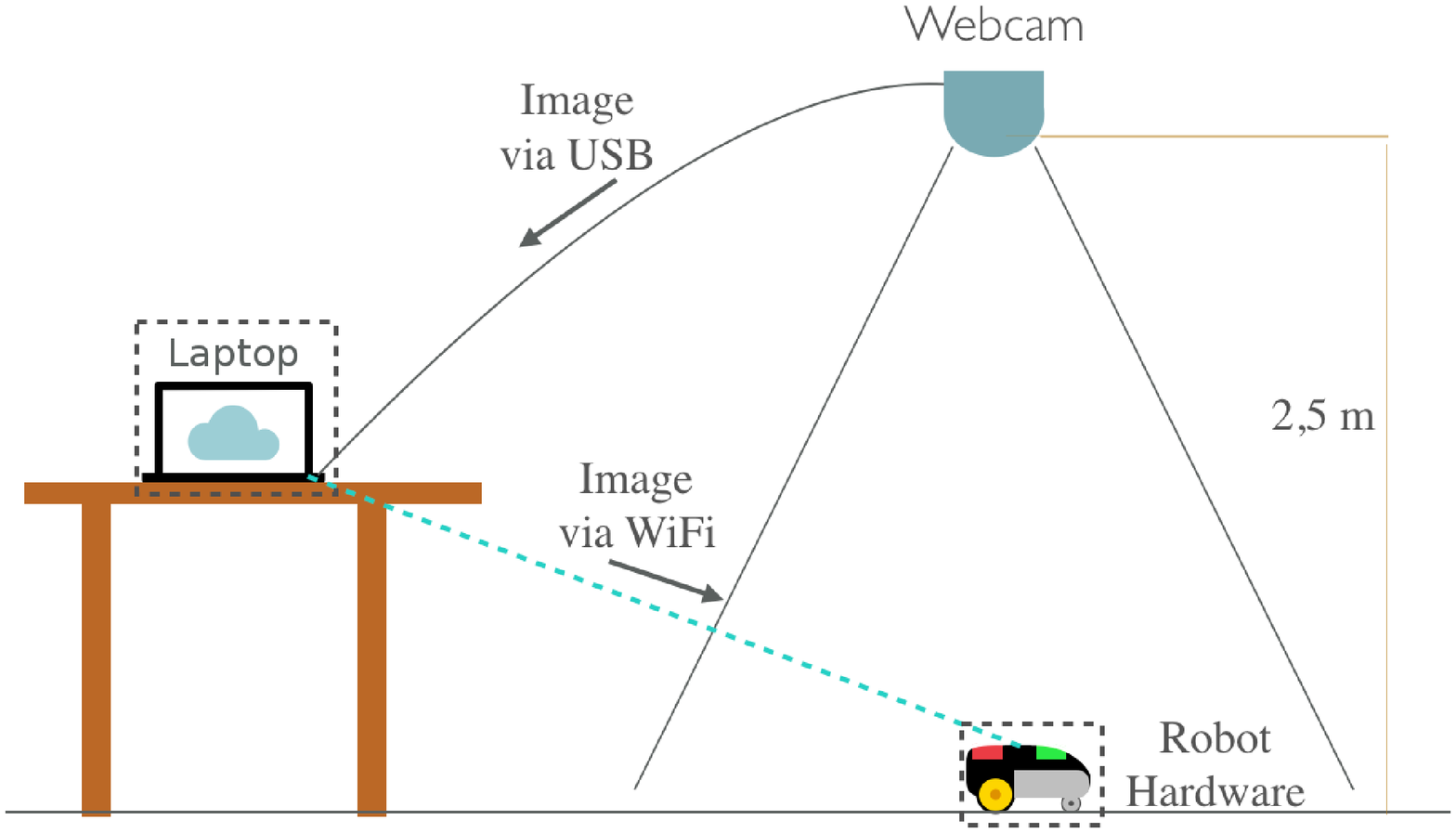}
	  \label{fig:setup}
	\end{minipage}}
 \hfill 
  \subfloat[]{
	\begin{minipage}[c][1.1\width]{
	   0.33\textwidth}
	   \centering
	  \includegraphics[width=0.7\linewidth]{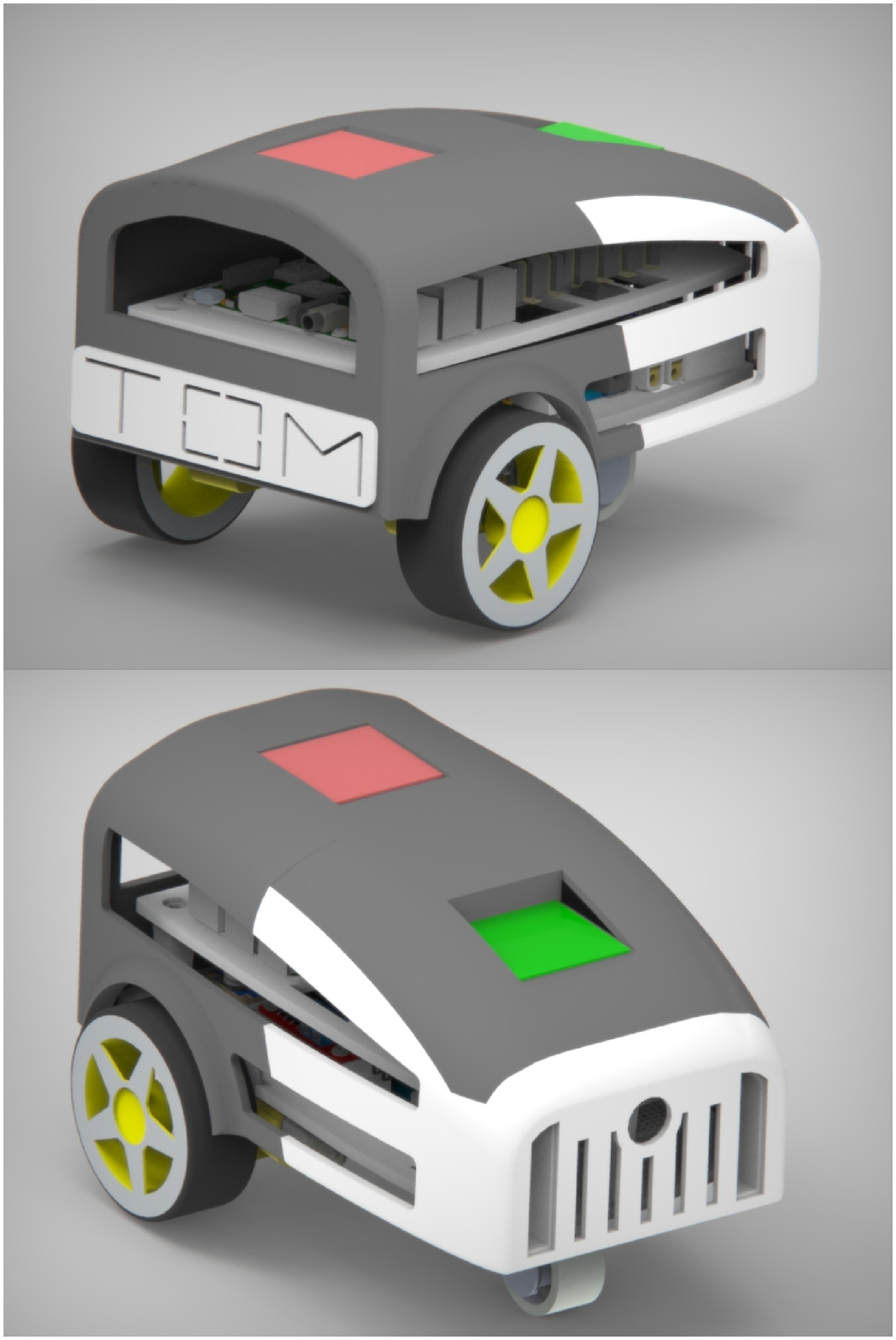}
	  \label{fig:robot}
	\end{minipage}}
 \hfill
\caption{(a) Overview of the ceiling mounted camera and image transmission, (b) 
Color tags mounted on the robot for positioning and orientation.}
\end{figure}

In Fig.~\ref{fig:conexion} we show an overview for our \arch{Robot Hardware} and 
\arch{Sensor} solution. We selected a simple two-wheeled solution as the robot platform, 
similar to~\cite{twowheeljaiio}, where we use a third pivoting 
wheel to have a stable platform (see Fig.~\ref{fig:render}) and avoid incurring in 
self-balancing control algorithms as in~\cite{twowheel2}. We defined the size and shape 
of the robot to be able to accommodate all the electronic and hardware components 
and have easy access to the battery.

A simplified connection schematics for the robot design is shown in 
Fig.~\ref{fig:conexion}. Due to the motion control algorithm defined in 
Section~\ref{sec:hybrid} we require independent motors to drive the wheels, for which we 
selected common DC motors and an H-bridge to command them. An Arduino UNO 
board was used to communicate with the H-bridge, and to later implement the low-level 
feedback- and motion control logic. For synthesising and enacting the discrete event 
controllers, and implementing the image processing modules we included a Raspberry Pi 3B+ 
single-board computer. 
For power we used 6 AA rechargeable batteries (with a voltage regulator for the 
Raspberry Pi), generating an autonomy of over 1 hour. 
Finally, for the alarm capability we connected to the Arduino leds and a speaker to 
produce audible noise (not shown in Fig.~\ref{fig:conexion}).

Our motion control algorithm requires to sense both the position and heading of the 
robot, as well as automatic detection of static obstacles for the planning layer. Since 
we are planning to do missions 
on the footprint of around $\SI{2}{\metre} \times \SI{3}{\metre}$ (e.g., the floor of a 
small room), we used a camera mounted on the ceiling to track the movement. We added color 
tags on top of the robot (see Fig.~\ref{fig:robot}) 
and used blue obstacles to automatically detect from the images: position and orientation 
of the robot and discrete regions that include obstacles. The images from the 
ceiling-cam are transmitted via USB to an intermediary laptop and then via WiFi to the 
onboard computer (see Fig.~\ref{fig:setup}), to be later processed onboard.

\subsection{Feedback-controllers and Direct Actuators} \label{sec:actuator}

For the alert system we implemented the following \arch{Direct Actuator}: a 
function on the Arduino that turns on the leds and the beeping noise depending on the 
value of a input flag from the serial port. 

For the motion capabilities we used a \arch{Feedback-controller} for orientation and 
an open-loop controller for forward movement. The latter is fairly straightforward to 
implement if we don't want a controlled velocity: we implemented at the Arduino level a 
function to set the wheels into a forward movement through a constant PWM signal for both 
wheels. This controller does not guarantee that the robot 
will follow a straight path given that, due to slipping or friction, a tendency to turn 
can be present. We simply ignore this problem and expect the orientation controller to 
fix the heading of the vehicle as it moves.

Based on \cite{franklin}, to design the feedback-controller we must first develop a 
model of the dynamics of the system. In \cite{franklin}, a 
mathematical model is proposed with a simplified electrical circuit and mechanical forces.
Thus, we obtained the dynamical model (Laplace transformed) of 
the system shown in (\ref{eqn:system}), where we group the many circuit and 
mechanical related constants ($r_w, K_t, J_m, l, K_e, K_t, J_m$) into $K$ and $D$, 
$\theta$ is the orientation of the robot and $V$ is the applied voltage to the motors 
(with different signs for each one).
\begin{equation} \label{eqn:system}
\frac{\theta (s)}{V(s)} = \frac{r_{w} \frac{K_t}{J_m l}}{s (s+\frac{K_e K_t}{J_m})} = 
\frac{K}{s(s+D)}
\end{equation}

\begin{figure}[bt]
\centering
\hfill
  \subfloat[]{
	\begin{minipage}[c][0.3\width]{
	   0.47\textwidth}
	   \centering
	  \includegraphics[width=1\linewidth]{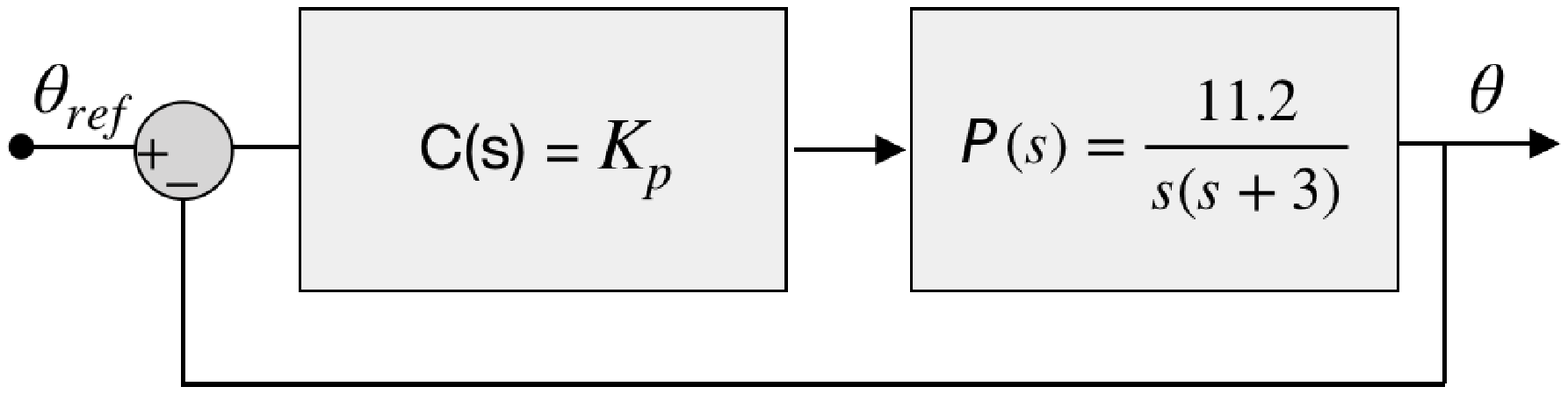}
	  \label{fig:controlloop}
	\end{minipage}}
 \hfill 
  \subfloat[]{
	\begin{minipage}[c][0.3\width]{
	   0.47\textwidth}
	   \centering
	  \includegraphics[width=1\linewidth]{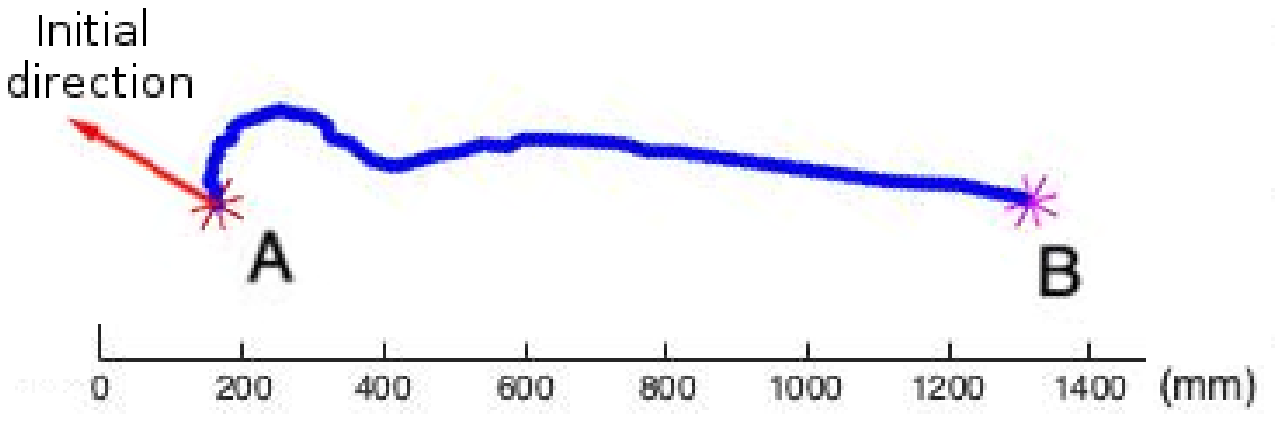}
	  \label{fig:mpath}
	\end{minipage}}
	\hspace{6mm}
\caption{(a) Feedback-controller showing the identified system parameters, (b) 
Example path generated by the motion controller going from a location $A$ to $B$.}
\end{figure}

Fig.~\ref{fig:controlloop} shows the identified parameters $K$ and $D$ using 
standard identification control techniques. This system can be reasonably controlled 
using a proportional controller as shown also in Fig.~\ref{fig:controlloop}, where $K_p = 
3.12$.

\subsection{Hybrid Control Layer Implementation}

As recognised in~\cite{Castro15}, discretization for synthesis problems must take into 
account the actual ability of the robot to transverse from a discrete cell to the 
next without invading nearby cells in the process. The main limitation of the 
motion control strategy is at the feedback-control and hardware level, where from 
experimental data we estimated that the robot requires 15 pixels (i.e., 
\SI{100}{\milli\metre}) of turn radius to be able to rotate, mainly because it 
doesn't properly rotate around its centre of mass. This makes us set the cell size for the 
discretization at 60 pixels (i.e., \SI{400}{\milli\metre}), and as result gives us a 
rectangular grid of $4 \times 5$ cells for the workspace defined by the mounted camera 
(see Fig.~\ref{fig:setup}). 

The final step of our design methodology is implementing the different modules in the 
\arch{Hybrid Control Layer}. The developed modules were implemented in Python 3.4 onboard 
the vehicle and behave as follows:

\textbf{Alert Module}: This module interprets the \alerton and \alertoff commands 
and sets the flag described in Section~\ref{sec:actuator} accordingly.

\textbf{Motion Handler Module}: The motion handler takes as an input 
\go{i}{j} and translates it into a continuous $(x,y)$ position (in units of distance) 
to later command the \arch{Motion Control} component. It also detects from the available 
\arch{sensor data} that the target discrete location has been reached, i.e., the robot's 
position is less than 15 pixels (i.e., \SI{100}{\milli\metre}) from the centre of 
the discrete location, and outputs the discrete event \arrived{i}{j} to the 
\arch{Enactor}.
 
\textbf{Motion Control}: This module implements the algorithm described in 
Section~\ref{sec:hybrid}, generating trajectories for the robot as shown in 
Fig.~\ref{fig:mpath}. Since the forward movement is generated through open-loop 
control, this module also switches between the rotating and forward motion capabilities, 
to keep the robot pointing to the target location. Part of this logic is 
implemented on the Arduino.

\section{Experimentation and Validation} \label{sec:validation}

\begin{figure}[bt]
\centering
\hfill
  \subfloat[]{
	\begin{minipage}[c][0.7\width]{
	   0.32\textwidth}
	   \centering
	  \includegraphics[width=1\linewidth]{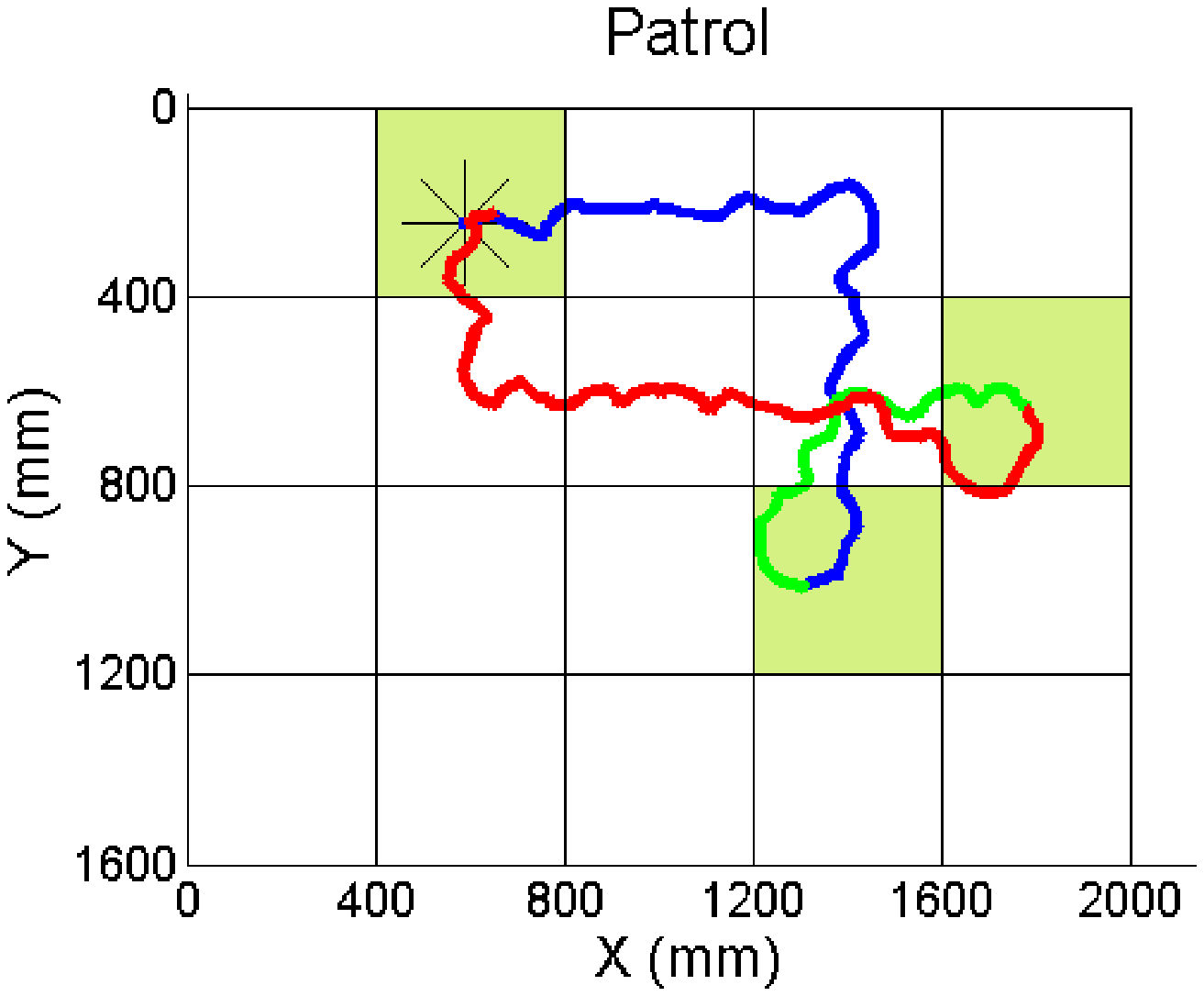}
	  \label{fig:patrol}
	\end{minipage}}
 \hfill 
  \subfloat[]{
	\begin{minipage}[c][0.7\width]{
	   0.32\textwidth}
	   \centering
	  \includegraphics[width=1\linewidth]{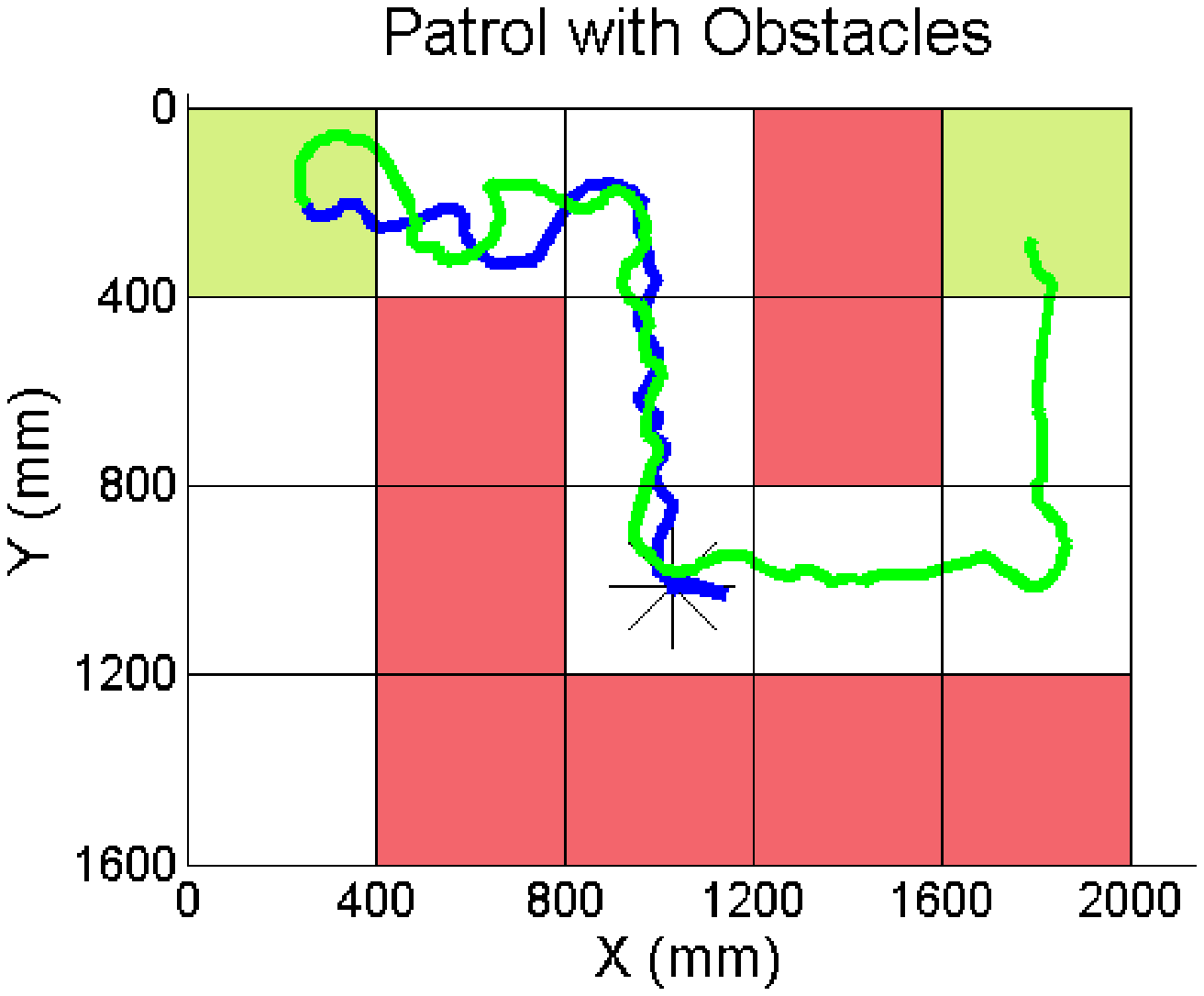}
	  \label{fig:obstacles}
	\end{minipage}}
 \hfill
  \subfloat[]{
	\begin{minipage}[c][0.7\width]{
	   0.32\textwidth}
	   \centering
	  \includegraphics[width=1\linewidth]{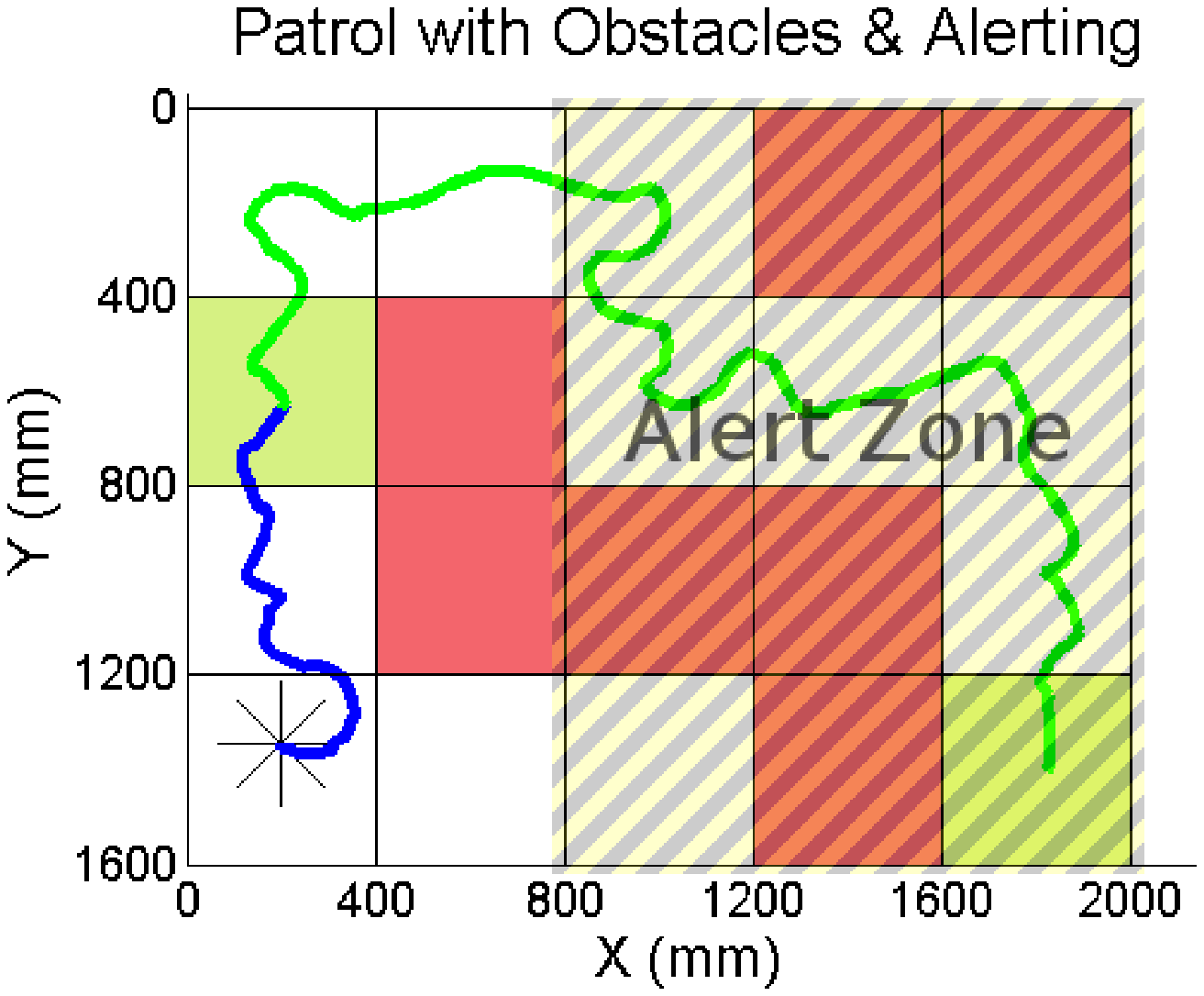}
	  \label{fig:alert}
	\end{minipage}}
 \hfill
\caption{(a) Patrol of three locations, (b) 
Patrol of two locations with obstacles, (c) Patrol of two locations with obstacles and 
alerting. \textit{Note:} Patrol locations are shown in green, obstacles in red and alert 
zones with stripes.}
\end{figure}

We validated the correct behaviour of the designed robot by testing it in three different 
missions taken from its domain of application (see Section~\ref{sec:abstraction}). The 
videos and the full FLTL specifications from which the discrete event controllers were 
synthesised can be found in the supplementary material.

Our first validation mission was a simple surveillance of three areas. For this we 
selected three areas to visit infinitely often (see Fig.~\ref{fig:patrol}) and used a 
similar specification to $\varphi_1$ in equation (\ref{eqn:sp_ex}) adjusted to these 
three locations. This mission allowed us to test the correct behaviour of the overall 
system and analyse the paths generated by the motion control strategy. We run the 
synthesised controller on the vehicle, and it generated as a result the trajectory shown 
in Fig.~\ref{fig:patrol}, successfully doing a full patrol loop around the three 
locations.

For our second mission we chose to validate our automatic detection of obstacles. We 
chose for this a patrol mission of two areas as shown in Fig.~\ref{fig:obstacles} and 
included in the workspace several blue objects, that we can automatically detect with 
simple image filtering techniques and tag them as locations to avoid with a similar 
specification as $\varphi_2$ in equation (\ref{eqn:sp_ex}). Fig.~\ref{fig:obstacles} 
shows that the robot successfully visits the two locations while avoiding the red areas.

Finally, we tested the alert system by synthesising a mission similar to the last one but 
turning on the alarm sounds and leds while in the regions shown in 
Fig.~\ref{fig:alert}, using similar specification as $\varphi_3$ of (\ref{eqn:sp_ex}). 
Fig.~\ref{fig:alert} shows the resulting path for this 
mission. Note that although the path followed by the robot is quite curvy at times, it 
successfully accomplishes the mission it was designed for.
\section{Conclusions and Future Work} \label{sec:conclusions}

In this work we presented a design methodology and effectively used it to build a 
robot system that successfully accomplishes a set of missions taken from a surveillance 
domain of application. For this, we built on discrete event controller synthesis techniques 
which allowed us to generate plans that are guaranteed to accomplish high-level mission 
specifications if a number of assumptions about the robot's capabilities and workspace 
hold. We validated our system by running three increasingly complicated tasks that 
involved both motion, obstacle avoidance and the non-movement functionality of an alarm 
system.

In future work it would be interesting to incorporate into the domain of 
application the use of more complex actuators (e.g., manipulating capabilities) 
to see if the presented methodology is helpful in these scenarios. Another aspect is to 
focus on improving the motion paths described by the robot, analysing the balance between 
more sophisticated motion control/planning, more robust feedback-controllers and 
different hardware/mechanical designs.

%
%
\bibliographystyle{plain}
\bibliography{paper}

\end{document}